\documentclass[lettersize,journal]{IEEEtran}
\usepackage{amsmath,amsfonts}
\usepackage{algorithmic}
\usepackage{algorithm}
\usepackage{array}
\usepackage[caption=false,font=normalsize,labelfont=sf,textfont=sf]{subfig}
\usepackage{textcomp}
\usepackage{stfloats}
\usepackage{url}
\usepackage{verbatim}
\usepackage{graphicx}
\usepackage{cite}
\usepackage{tikz}
\usepackage{pgfplots}
\usepackage{makecell}
\begin{document}

\title{Depression Detection in Social Media Posts Using Transformer-Based Models and Auxiliary Features}

\author{Marios Kerasiotis, Loukas Ilias, Dimitris Askounis
\thanks{The authors are with the Decision Support Systems Laboratory, School of Electrical and Computer Engineering, National Technical University of Athens, 15780 Athens, Greece (e-mail: marioskerasiotis@gmail.com; lilias@epu.ntua.gr; askous@epu.ntua.gr).}}



\maketitle

\begin{abstract}
The detection of depression in social media posts is crucial due to the increasing prevalence of mental health issues. Traditional machine learning algorithms often fail to capture intricate textual patterns, limiting their effectiveness in identifying depression.
Existing studies have explored various approaches to this problem but often fall short in terms of accuracy and robustness. To address these limitations, this research proposes a neural network architecture leveraging transformer-based models combined with metadata and linguistic markers.
The study employs DistilBERT, extracting information from the last four layers of the transformer, applying learned weights, and averaging them to create a rich representation of the input text. This representation, augmented by metadata and linguistic markers, enhances the model’s comprehension of each post. Dropout layers prevent overfitting, and a Multilayer Perceptron (MLP) is used for final classification.
Data augmentation techniques, inspired by the Easy Data Augmentation (EDA) methods, are also employed to improve model performance. Using BERT, random insertion and substitution of phrases generate additional training data, focusing on balancing the dataset by augmenting underrepresented classes.
The proposed model achieves weighted Precision, Recall, and F1-scores of 84.26\%, 84.18\%, and 84.15\%, respectively. The augmentation techniques significantly enhance model performance, increasing the weighted F1-score from 72.59\% to 84.15\%.
\end{abstract}


\section{Introduction}
\label{sec:introduction}
According to the World Health Organization (WHO), a mental disorder is characterized by a clinically significant disturbance in an individual’s cognition, emotional regulation, or behaviour \cite{world_health_organization_mental_2023}. WHO's data from 2019 indicated that 1 in 8 people suffer from mental disorders. However, the outbreak of the COVID-19 pandemic brought unprecedented challenges and led to a significant increase in the prevalence of mental health issues. Specifically, anxiety disorders surged by 26\%, and depression cases rose by 28\% \cite{world_health_organization_mental_2023}. Among the various types of mental disorders, depression emerged as a major concern. It is estimated that 280 million people suffer from depression, contributing to a distressing number of over 700 thousand suicides \cite{world_health_organization_depressive_2023}. In the modern digital age, social media platforms have become popular outlets for individuals to share their emotions and experiences. Many of these users are affected by mental health disorders, and researchers have recognized the potential of social media data to reveal valuable insights and linguistic patterns related to mental health. By analyzing the vast quantity of information available on social media, researchers aim to better understand mental health issues and their prevalence.

In recent years, the field of depression detection in posts has witnessed significant advancements with the emergence of state-of-the-art machine learning models, showing promise in identifying potential signs of depression from social media posts. However, these models have notable limitations. Firstly, many rely on traditional machine learning algorithms such as Logistic Regression (LR) \cite{eichstaedt_facebook_2018,huang_suicidal_2019}, Support Vector Machines (SVM) \cite{leiva_towards_2017,aldarwish_predicting_2017,islam_depression_2018}, Decision Trees \cite{leiva_towards_2017,uban_emotion_2021,ameer_mental_2022}, Random Forest (RF) \cite{kamite_detection_2020,leiva_towards_2017,cacheda_early_2019}, AdaBoost \cite{chiong_textual-based_2021,safa_automatic_2022}, and k-Nearest Neighbors (k-NN) \cite{islam_depression_2018,leiva_towards_2017}, which, while useful, may lack the sophistication needed to fully grasp the complexities of mental health-related textual data, leading to suboptimal performance. At the same time, extracting a large number of features is a time-consuming process demanding a level of domain expertise. The optimal set of features may not be found. Additionally, the training of traditional ML algorithms usually achieves suboptimal performance, while these algorithms present issues of generalization to new data. To address this limitation, researchers have turned to more advanced techniques, including deep learning approaches like Long Short-Term Memory (LSTM) \cite{shah_early_2020,wu_deep_2020,cong_x--bilstm_2018}, Convolutional Neural Networks (CNN) \cite{yao_detection_2020,trotzek_utilizing_2020,yates_depression_2017,gkotsis_characterisation_2017} , and then transformers \cite{chen_detecting_2023,uban_emotion_2021,lin_sensemood_2020,rao_mgl-cnn_2020,kabir_deptweet_2023,murarka_detection_2020} like BERT \cite{DBLP:journals/corr/abs-1810-04805}. These deep learning methods have shown promise in capturing intricate patterns within textual content and achieving higher levels of contextual understanding. However, simpler neural networks that do not employ transformers may not fully capture these intricate patterns and consequently may not perform as effectively as the transformer-based models, which are pre-trained with vast corpora \cite{wikidump,Zhu_2015_ICCV}. On the other hand, many researchers have incorporated transformers into their work, but they often neglect the use of other valuable metadata and linguistic markers that could potentially enhance the classification of individuals as either depressed or not.

To overcome the limitations of traditional machine learning algorithms and to leverage the strengths of transformer-based models, this research introduces a novel neural network architecture tailored for depression detection in social media posts. The proposed model builds upon the power of distilbert-base-uncased \cite{Sanh2019DistilBERTAD}, a widely used transformer model, by extracting information from the last four layers. These layers are then multiplied by learned weights and averaged, resulting in a rich representation that captures essential linguistic patterns in the input text. The averaged CLS (classification) tokens from BERT, serving as powerful contextual embeddings, are then concatenated with additional relevant metadata and linguistic markers. This strategic fusion of contextual embeddings and auxiliary features enhances the model's ability to grasp a holistic understanding of each post, enabling it to identify potential signs of depression more accurately. To further improve the generalization and robustness of the model, dropout layers are introduced to prevent overfitting during training. After incorporating dropout, the output is passed through a Multi-Layer Perceptron (MLP), which serves as the classification head. The MLP is designed to make the final decision regarding the post's depressive status, providing a multi-label classification output. By combining the transformer-based contextual embeddings with carefully selected features and incorporating dropout regularization, the proposed neural network achieves a higher level of sophistication compared to traditional algorithms while avoiding overreliance on vast pre-training corpora. This allows the model to effectively address the intricacies of mental health-related textual data, leading to improved performance in identifying depressive posts on social media platforms. The integration of additional metadata and linguistic markers also contributes to the model's ability to contextualize each post, capturing vital nuances that might otherwise be overlooked, thus making a valuable contribution to the field of depression detection in social media content. 

Our main contributions can be summarized as follows:

\begin{itemize}
    \item We propose a data augmentation approach to deal with the imbalanced dataset and improve the performance of the introduced model.  
    \item We present a new method for recognizing the depression severity level, which enables the fusion of contextual embeddings from BERT with valuable metadata and linguistic markers. 
    \item We introduce a weighted average approach to extract hidden states from various transformer layers, effectively capturing both simple and complex semantic information.
    \item We conduct a series of ablation experiments to prove the effectiveness of the proposed architecture.
\end{itemize}

\section{Related Work}
\label{sec:related}
The landscape of machine learning and natural language processing research is continuously evolving, and a variety of techniques have been employed to detect signs of depression in social media posts. Researchers have leveraged data from various sources including Twitter, Facebook, Reddit, and Instagram, applying machine learning algorithms to analyze these vast datasets. The rich body of existing literature manifests the diverse approaches and methodologies undertaken to detect depressive symptoms through textual analysis.

\subsection{Works that use traditional machine learning algorithms}

The paper by Aragón et al. \cite{aragon_detecting_2023} proposes a novel approach to detect mental disorders in social media users based on their fine-grained emotions and temporal patterns. They compare their approach with previous works that have used linguistic, sentiment, and emotion features for the same task. They also consider deep learning models with attention mechanisms that can capture the temporal changes of emotions over time. Their approach combines two representations based on sub-emotions, which are clusters of words that express more specific and subtle emotions. They use a fusion strategy to integrate the presence and variability of sub-emotions in the users’ posts. Their approach achieves better results than the baselines and provides some interpretability by highlighting the relevant sub-emotions for each mental disorder. They also contrast the emotional patterns of depression and anorexia, finding some distinctive features.

The paper by Guntuku et al. \cite{guntuku_understanding_2019} explores the use of social media language to measure and understand psychological stress. They collect Facebook and Twitter data from 601 users who also completed a stress questionnaire. They extract linguistic, sentiment, emotion, and topic features from the social media posts and use them to predict stress scores. They also use LIWC, a psycholinguistic dictionary, and Tensistrength, a tool to detect stress expressions, to compare the language of stressed and non-stressed users. They find that Facebook language is more predictive of stress than Twitter language, and that fine-grained emotions and temporal patterns are important indicators of stress. They also apply domain adaptation techniques to transfer the user-level models to county-level Twitter language and validate them against survey-based stress measurements. They show that language-based stress estimates correlate with health behaviors and socioeconomic characteristics of counties.

The paper by Cacheda et al. \cite{cacheda_early_2019} proposes two machine learning methods for early detection of depression based on social media posts. They use a dataset collected from Reddit, which contains writings from users who have self-reported depression diagnosis and a control group. They extract textual, semantic, and writing features such as the number of words, time gap between posts, and text similarity from the social media posts and use them to train two random forest classifiers: one for detecting depressed subjects and another for identifying non-depressed subjects. They evaluate their methods using a time-aware metric that rewards early detections and penalizes late detections. They show that the dual model outperforms the singleton model and the state-of-the-art baselines by more than 10\%.

The paper by Coppersmith et al. \cite{coppersmith_quantifying_2014} explores the use of Twitter language to quantify mental health signals for four disorders: depression, bipolar, PTSD, and SAD. They use a novel method to collect data from users who self-report their diagnoses, and extract features based on LIWC, language models, and pattern of life analytics. They show that statistical classifiers can differentiate users with mental health disorders from control users, and that there are linguistic and behavioral differences among the disorders. They also conduct a correlation analysis to reveal the relationships between the features and the disorders. They suggest that Twitter data can provide a rich source of information for mental health research at both individual and population levels.

\subsection{Works that use neural networks}

The paper by Tadesse et al. \cite{tadesse_detection_2019} proposes a novel approach to detect depression in Reddit social media posts based on fine-grained emotions and temporal patterns. They use a dataset of 1841 posts collected from subreddits related to depression and non-depression. They extract features based on N-grams, LIWC, and LDA from the posts and use them to train five classification models: Logistic Regression, Support Vector Machine, Random Forest, Adaptive Boosting, and Multilayer Perceptron. They show that the combination of LIWC, LDA, and bigram features with the Multilayer Perceptron classifier achieves the best performance for depression detection with 91\% accuracy and 93\% F1 score. They also analyze the linguistic and emotional differences between depressed and non-depressed users.

The paper by Chiong et al. \cite{chiong_textual-based_2021} proposes a textual-based featuring approach for depression detection using machine learning classifiers and social media texts. They use two public Twitter datasets and three non-Twitter datasets to train and test various single and ensemble classifiers, such as logistic regression, support vector machine, multilayer perceptron, random forest, and adaptive boosting. They also apply dynamic sampling methods to deal with imbalanced data. They show that their approach can effectively detect depression in social media texts even without relying on specific keywords, such as ‘depression’ and ‘diagnose’. They also demonstrate the generality of their approach by applying it to different social media sources, such as Facebook, Reddit, and an electronic diary.

The paper by Yao et al. \cite{yao_detection_2020} investigates the detection of suicidality among opioid users on Reddit using machine learning methods. They use data from subreddits related to suicide, depression, opioid abuse, and control topics, and extract features based on TF-IDF, word embeddings, and character embeddings. They compare several traditional and neural network classifiers, such as logistic regression, random forest, support vector machine, FastText, recurrent neural network, attention-based bidirectional recurrent neural network, and convolutional neural network. They show that convolutional neural network achieves the best performance for both tasks: distinguishing between suicidal and non-suicidal language, and detecting opioid addiction among suicidal posts. They also use Amazon Mechanical Turk to annotate out-of-sample data and evaluate the prediction ability of the models. They conclude that social media platforms such as Reddit can provide valuable information for mental health research and monitoring.

The paper by Trotzek et al. \cite{trotzek_utilizing_2020} presents a novel approach to detect depression in social media users based on linguistic metadata and neural networks. They use the eRisk 2017 dataset, which contains chronological sequences of posts and comments from Reddit users who self-reported depression diagnosis and a control group. They extract 27 features based on word and grammar usage, readability, and specific phrases from the text content of the users and use them to train various machine learning classifiers. They also use word embeddings based on word2vec, fastText, and GloVe to vectorize the text content and feed it to convolutional neural networks. They show that the combination of linguistic metadata and convolutional neural networks achieves the best performance for early detection of depression with 71\% F1 score and 12.13\% ERDE5 score. They also propose a modification of the ERDE metric to make it more intuitive and adaptable.

The paper by Chen and Wang \cite{chen_advanced_2018} proposes an advanced model for Twitter sentiment analysis based on the LSTM-CNN model presented by Sosa \cite{sosa_twitter_nodate}. The authors combine the encoder-decoder framework with the multi-layer LSTM-CNN model, in which LSTM can 'remember' forward information of the sequence and multi-layer CNN can capture and learn local information effectively. The encoder-decoder part is used to reconstruct the input matrix, making the feature extraction and learning more intrinsic and effective. The authors compare their model with single-layer CNN, multi-layer CNN, LSTM-CNN, and CNN-LSTM models on a combined dataset of tweets labeled as 'positive' or 'negative'. The results show that their model achieves the state-of-the-art accuracy of 78.6\%, outperforming all the other models. The authors also analyze the effect of different parameters and layers on the performance of their model. The paper acknowledges some limitations of the model, such as the difficulty of handling sarcasm, irony, and humor, and the lack of external knowledge sources, such as world knowledge and common sense. Lastly it suggests some directions for future work, such as connecting the model with Part-of-Speech (POS) tagging.

 Anshul et al. \cite{10241281} proposed a multimodal framework for depression detection during Covid-19 using social media data, integrating textual, user-specific, and image analysis. They extracted features from tweet content, URLs, and images, and utilized a Visual Neural Network (VNN) for image embeddings. Their model achieved the best results with a precision of 93.3\%, recall of 90.6\%, F1-score of 91.9\%, and accuracy of 91.7\%. Limitations include the model's reliance on labeled data, potential bias due to the dataset's specific demographic, and the challenge of accurately interpreting multimodal data. Future work will focus on enhancing model generalization across diverse populations, incorporating additional data sources for robustness, and improving interpretability through explainable AI techniques.

\subsection{Works that use transformers}
Transformer models, including BERT and its variants such as RoBERTa\cite{DBLP:journals/corr/abs-1907-11692} and DistilBERT\cite{Sanh2019DistilBERTAD}, are playing a pivotal role in the field of mental health detection on social media, as demonstrated by a variety of complex academic papers.

The paper by Uban et al. \cite{uban_emotion_2021} presents a comprehensive study of mental disorders in social media, from different perspectives. They use three datasets from the eRisk Lab, which contain Reddit posts from users diagnosed with depression, anorexia, or self-harm. They extract features based on content, style, emotions, and cognitive styles from the posts and use them to train various deep learning models, such as bidirectional LSTM, CNN, hierarchical attention network, and transformers. They show that the hierarchical attention network with LSTM encoders achieves the best performance for early detection of mental disorders, with high F1 and AUC scores. They also interpret the behavior of their models by analyzing the attention weights, the feature importance, and the correlation between emotions and cognitive styles. They reveal some distinctive linguistic and emotional patterns for each mental disorder and provide insights for mental health research.

The paper by Lin et al. \cite{lin_sensemood_2020} presents SenseMood, a system for depression detection on social media using multimodal data. They use a Twitter dataset that contains posts from users who self-reported depression diagnosis and a control group. They extract features based on CNN and BERT from the images and texts posted by the users and use them to train a neural network classifier. They show that their system can achieve better performance than several baselines and provide an analysis report for each user. They also discuss the challenges and opportunities of using social media for mental health research and monitoring.

The paper by Rao et al. \cite{rao_mgl-cnn_2020} proposes two hierarchical posts representations models for identifying depressed individuals in online forums, namely MGL-CNN and SGL-CNN. The models use convolutional neural networks with gated units to learn the key features of the posts and the users' emotional states. In addition to the MGL-CNN and SGL-CNN models, the paper also introduces a novel approach that combines BERT with LSTM. This hybrid model leverages the strengths of BERT in understanding the context of words in sentences and LSTM’s ability to remember patterns over time. This combination allows the model to capture both the semantic meaning of the posts and the temporal patterns of the users’ emotional states. The models are evaluated on two datasets: the Reddit Self-reported Depression Diagnosis dataset and the Early Detection of Depression (eRisk) dataset. The results show that the models outperform several baselines and state-of-the-art methods in terms of precision, recall, and F1-score. The models also demonstrate robustness and generality across different online forums. The paper discusses the limitations of the models, such as the difficulty of capturing long-term dependency and the lack of external knowledge. For future work, the authors will explore the application of MGL-CNN and SGL-CNN to general document-level sentiment analysis.

The paper by Kabir et al. \cite{kabir_deptweet_2023} presents DEPTWEET, a new dataset for depression detection on social media using fine-grained emotions and temporal patterns. They use a typology based on the PHQ-9 questionnaire and the mood scale of BipolarUK to label tweets as non-depressed, mildly depressed, moderately depressed, or severely depressed. They also provide a confidence score for each label to reflect the annotator agreement. They collect 40191 tweets from Twitter using 88 depression-related keywords and annotate them with the help of expert psychologists. They compare four baseline models: SVM, BiLSTM, BERT, and DistilBERT, and show that DistilBERT achieves the best performance with 82\% F1 score. They also analyze the linguistic and emotional features of different depression severities.

The paper by Murarka et al. \cite{murarka_detection_2020} proposes a RoBERTa-based classifier for detecting and classifying mental illnesses on social media using fine-grained emotions and temporal patterns. They use a new dataset collected from Reddit, which contains posts from users who self-reported depression, anxiety, bipolar disorder, ADHD, or PTSD diagnosis and a control group. They use LSTM, BERT and RoBERTa and they  show that their system can achieve better performance than several baselines and provide an analysis report for each user. They also discuss the challenges and opportunities of using social media for mental health research and monitoring and as part of their future work they plan to collaborate with professionals to annotate their dataset.

The paper by Wang et al. \cite{wang_depression_2020} explores the potential of deep-learning methods with pretrained language representation models for depression risk prediction from Chinese microblogs. The authors use a manually annotated dataset of 13993 microblogs collected from Sina Weibo, and compare three deep-learning methods: BERT, RoBERTa, and XLNET. They also investigate the effect of further pretraining the language models on a large-scale unlabeled corpus from Weibo. The results show that the deep-learning methods outperform previous methods, and that further pretraining leads to better performance. The authors acknowledge the limitations of their dataset, such as data imbalance and ambiguous words, and suggest possible directions for future work, such as expanding the dataset, using user-level context, and incorporating medical knowledge. The paper contributes to the field of depression health care by demonstrating the feasibility and effectiveness of using social media data and deep-learning methods to discover potential patients with depression and to trace their mental health conditions.

The paper by Malviya et al. \cite{malviya_transformers_2021} presents a novel approach for detecting depression in social media users using transformer models. They use a dataset collected from Reddit, which contains posts from users who self-reported depression diagnosis and a control group. The authors extract features based on TF-IDF, word embeddings, and character embeddings from the posts and use them to train various classifiers, including Support Vector Machines, Linear Classifiers, Bagging models, Boosting models, and Transformer models. They show that the transformer model achieves the highest accuracy (98\%) in detecting depression in social media users. The authors conclude that transformer models significantly outperform conventional models used for depression detection.

The paper by Ilias and Askounis \cite{ILIAS2023100270} introduces a novel method for detecting stress and depression in social media using multitask learning. The authors propose two architectures that use depression detection as the primary task and stress detection as the auxiliary task. The first architecture consists of a shared BERT layer and two task-specific BERT layers, while the second architecture adds an attention fusion network to weight the shared and task-specific representations. The authors use two different datasets for the primary and auxiliary tasks, collected under different conditions, to make the problem more challenging and realistic. The authors compare their approaches with state-of-the-art methods, single-task learning, and transfer learning, and show that their multitask learning frameworks achieve better performance for depression detection.

The paper by Haque et al. \cite{haque_transformer_2020} proposes a novel approach to detect suicidal ideation on social media using transformer models. They use a dataset of 3549 posts from Reddit, collected from the SuicideWatch subreddit and other non-suicidal subreddits. They extract features based on TF-IDF, word embeddings, and character embeddings from the posts and use them to train various classifiers, such as Bi-LSTM, BERT, ALBERT, ROBERTa, and XLNET. They show that ROBERTa outperforms all the other models with 95.21\% accuracy, 98.44\% recall, 92.67\% precision, and 95.47\% F1-score. They conclude that transformer models are more effective than conventional deep learning models for suicidal ideation detection.

The paper by Thushari et al. \cite{thushari2023identifying} focuses on identifying psychological well-being through Reddit posts from mental health groups like depression, anxiety, bipolar disorder, and SuicideWatch using the SWMH dataset. They employed the MentalBERT model, which outperformed other models with a 76.70\% accuracy. The study also highlights the importance of explainable AI, using LIME to validate the model's trustworthiness, which is crucial for understanding the complex interrelationships in mental health data.

The paper by Vajrobol et al. \cite{vajrobol2023explainable} addresses the challenge of detecting depression in low-resource languages, specifically focusing on Thai. To overcome the limited availability of annotated data and language models, the authors propose transferring knowledge from English to Thai. Their approach uses RoBERTa, which achieved the highest accuracy at 77.97\%, with strong recall, precision, and F1-scores. The study also emphasizes the importance of explainable NLP, demonstrating that RoBERTa effectively captures the context in both depression and non-depression classes. 

These studies demonstrate the importance of pre-processing, feature extraction, and the careful selection of machine learning algorithms, while also revealing ongoing limitations such as over fitting, small dataset sizes, annotator biases, and the lack of age and gender awareness. Future work in the field may continue to refine pre-processing techniques, mitigate dataset limitations, incorporate domain expertise, and investigate hybrid neural network architectures to enhance the efficacy of depression detection models.

\subsection{Related Work Review Findings}
In the context of the related literature, the limitations commonly observed across the existing methods include over fitting, small data set sizes, annotator biases, and a lack of age and gender awareness. Furthermore, some models tend to lean heavily on vast pre-training corpora, often leading to over-dependence on data and under performance when faced with unique or unfamiliar cases. Others might fall short of capturing all the nuances and contextual information necessary for the accurate detection of depressive symptoms. Many models may also overlook the crucial role that auxiliary features, such as metadata and linguistic markers, can play in enriching the understanding of the text. Consequently, while significant strides have been made in the field, there remains a need for a robust and comprehensive model that can overcome these limitations while effectively addressing the intricacies of mental health-related textual data.

The proposed model in this research offers several advancements and improvements over the methodologies reported in the literature. First, it leverages the power of BERT, a widely used transformer model, to extract detailed linguistic patterns from the input text. This transformer-based approach overcomes the limitations of traditional machine learning algorithms, offering a more sophisticated method for interpreting the complexities of natural language.  For example, shallow ML algorithms and deep neural networks often use GloVe, word2vec, or fastText embeddings. On the contrary, transformer-based networks, i.e., BERT, can capture the context of the input sequence effectively. Second, the model does not solely rely on transformer-based contextual embeddings. Instead, it incorporates auxiliary features - metadata and linguistic markers - into the analysis. This strategic fusion enhances the model's ability to comprehend each post holistically, capturing crucial nuances that might otherwise be overlooked. Consequently, this allows for a more accurate identification of potential signs of depression.

By overcoming the limitations of previous methodologies and introducing new elements such as auxiliary feature integration, the proposed model achieves a higher level of sophistication. It effectively addresses the challenges inherent in identifying depressive symptoms in social media content, leading to more accurate and reliable results. As such, this research makes a valuable contribution to the field of depression detection in social media posts.

\section{Dataset and preprocessing}
\label{sec:dataandprep}
In this study, we utilized the "Depression Severity Dataset" \cite{naseem_early_2022} for the task of early identification of depression severity levels on Reddit using ordinal classification. The dataset was collected and curated by Naseem et al. and was presented at the ACM Web Conference 2022. Researchers interested in utilizing the "Depression Severity Dataset" for their studies can access it from the GitHub repository \footnote{https://github.com/usmaann/Depression\_Severity\_Dataset}. The main objective of the dataset is to aid in the early identification of depression severity levels based on the textual content of Reddit posts. Each instance in the dataset consists of two columns:
\begin{enumerate}
    \item \textbf{Text:} This column contains the textual content of the Reddit posts. It serves as the input data for the classification task.
    \item \textbf{Label:} The label column represents the depression severity level associated with each Reddit post. The severity levels are categorized into four ordinal classes: Minimum (2587 posts), Mild (290 posts), Moderate (394), and Severe (282 posts).
\end{enumerate}

\subsection{Dataset augmentation}

To increase the size of the dataset and make it more robust for training our neural network, we applied data augmentation techniques. The goal was to generate additional samples from the existing dataset by perturbing the original text content without changing the underlying semantics. We focused on the Moderate, Mild, and Severe classes, as these classes had fewer instances compared to the Minimum class.

Firstly, we balanced the dataset to address the class imbalance issue. We sampled 250 instances from the Moderate class, 290 instances from the Mild class, and 281 instances from the Severe class. These samples were carefully selected to  ensure that the data are not merely repeated but instead varied through phrase changes to create a larger, more diverse dataset while maintaining a  balanced augmentation that does not overly equalize the data but keeps a proportional representation the depression severity levels.

In designing our approach to data augmentation, we took inspiration from the paper by Wei and  Zou \cite{wei-zou-2019-eda}. Their work introduced EDA, a set of four easy but powerful techniques: synonym replacement, random insertion, random swap, and random deletion. They demonstrated the effectiveness of these techniques in boosting performance, particularly for smaller datasets. Inspired by their work, we adapted and utilized a subset of these techniques, specifically insertion and substitution, to augment our selected samples.

To implement these techniques, we employed the pre-trained BERT (Bidirectional Encoder Representations from Transformers) model with the "bert-base-uncased" model \cite{DBLP:journals/corr/abs-1810-04805}. BERT is a powerful language model capable of generating contextual embeddings for words and sentences. In the insertion process, we randomly inserted new words into the text, leveraging BERT's language modeling capabilities to ensure the inserted words made contextual sense. In the substitution process, we replaced certain words with their contextual synonyms provided by BERT, mirroring the synonym replacement technique proposed by Wei and Zou.

After applying data augmentation to the selected samples, we observed changes in the distribution of depression severity levels. The proportions of the classes were modified as per Table \ref{table:Label_distribution}. As a result of the data augmentation process, the dataset now contains a total of 4,353 rows, with a more balanced representation of the different depression severity levels. This process is in line with the insights of the EDA paper, which emphasized the effectiveness of such augmentation methods, especially when dealing with imbalanced or smaller datasets. By adopting these principles, we believe our dataset is now more robust and suitable for training our neural network, aligning with the positive results demonstrated in the referenced work.

\begin{table}[!hbt]
    \centering
    \caption{Label distribution before and after text augmentation.}
    \label{table:Label_distribution}
    \begin{tabular}{|c|c|c|}
    \hline
        \textbf{Label} & \textbf{Before Augmentation (\%)} & \textbf{After Augmentation (\%)} \\ \hline
        Minimum & 72.68\% & 61.53\% \\ \hline
        Mild & 11.16\% & 14.24\% \\ \hline
        Moderate & 8.21\% & 12.22\% \\ \hline
        Severe & 7.96\% & 12.01\% \\ \hline
    \end{tabular}
\end{table}

\subsection{Dataset preparation}
Prior to utilizing the data for training our neural network, we conducted a comprehensive preprocessing routine to ensure the text was in a consistent and clean format. The preprocessing steps included converting all the text to lowercase to maintain uniformity, removing any HTML code present in the textual content, eliminating URLs and hyperlinks as they do not contribute to the semantics of the text, and removing subreddit mentions and user mentions to focus solely on the meaningful content. We also removed emojis, as they are graphical representations and not part of the natural language. Punctuation marks were removed to simplify the input data, and we resolved abbreviations and expanded contractions to ensure clarity and consistency. By performing these preprocessing steps, we obtained a clean and standardized version of the textual content from the dataset. This cleaned data will facilitate the training of our neural network, enabling it to focus on the meaningful patterns and linguistic features present in the text, ultimately leading to better performance in the task of classifying the severity of the posts.

\subsection{Feature extraction}
\label{subsec:feature_extr}
Feature extraction is a critical step in natural language processing, where we transform the preprocessed text into numerical representations that can be fed into machine learning models. In our study, we employed two different approaches for feature extraction: utilizing depression medication references and leveraging pre-trained emotion and sentiment models.

To incorporate potential markers of depression severity in the textual content, we performed feature extraction based on the presence of depression medication references. We curated a list of depression medications from a reputable source, Drugs.com\cite{httpswwwfacebookcomdrugscom_list_1970}, and searched for their occurrences in the posts. The idea behind this approach is that references to depression medication may indicate a higher likelihood of the posts belonging to the Moderate or Severe classes, as users might be discussing their treatment or experiences with such medications. This method of feature extraction allows us to leverage real-world data to potentially identify more severe cases of depression.

To capture the emotional content of the Reddit posts, we utilized the EmoRoBERTa \cite{kamath_enhanced_2022} model, which is a pre-trained language model specifically designed for emotion recognition. EmoRoBERTa can infer the emotions expressed in the text, such as happiness, sadness, fear, etc. By extracting emotions from the posts, we aimed to identify potential emotional patterns associated with different depression severity levels. For instance, an increased prevalence of negative emotions might be indicative of higher depression severity.

Sentiment analysis helps us gauge the overall sentiment of the Reddit posts as positive, negative, or neutral. For this purpose, we employed the CardiffNLP Twitter-RoBERTa model\cite{barbieri-etal-2020-tweeteval}, which is pre-trained on social media text and is well-suited for sentiment classification in informal language. The sentiment analysis enables us to understand the overall tone and emotional disposition of the posts and how it correlates with the different depression severity levels.

\section{Methodology}
\label{sec:method}
\subsection{Architecture}
\begin{figure*}[!htb]
    \centering
    \includegraphics[width=0.9\textwidth]{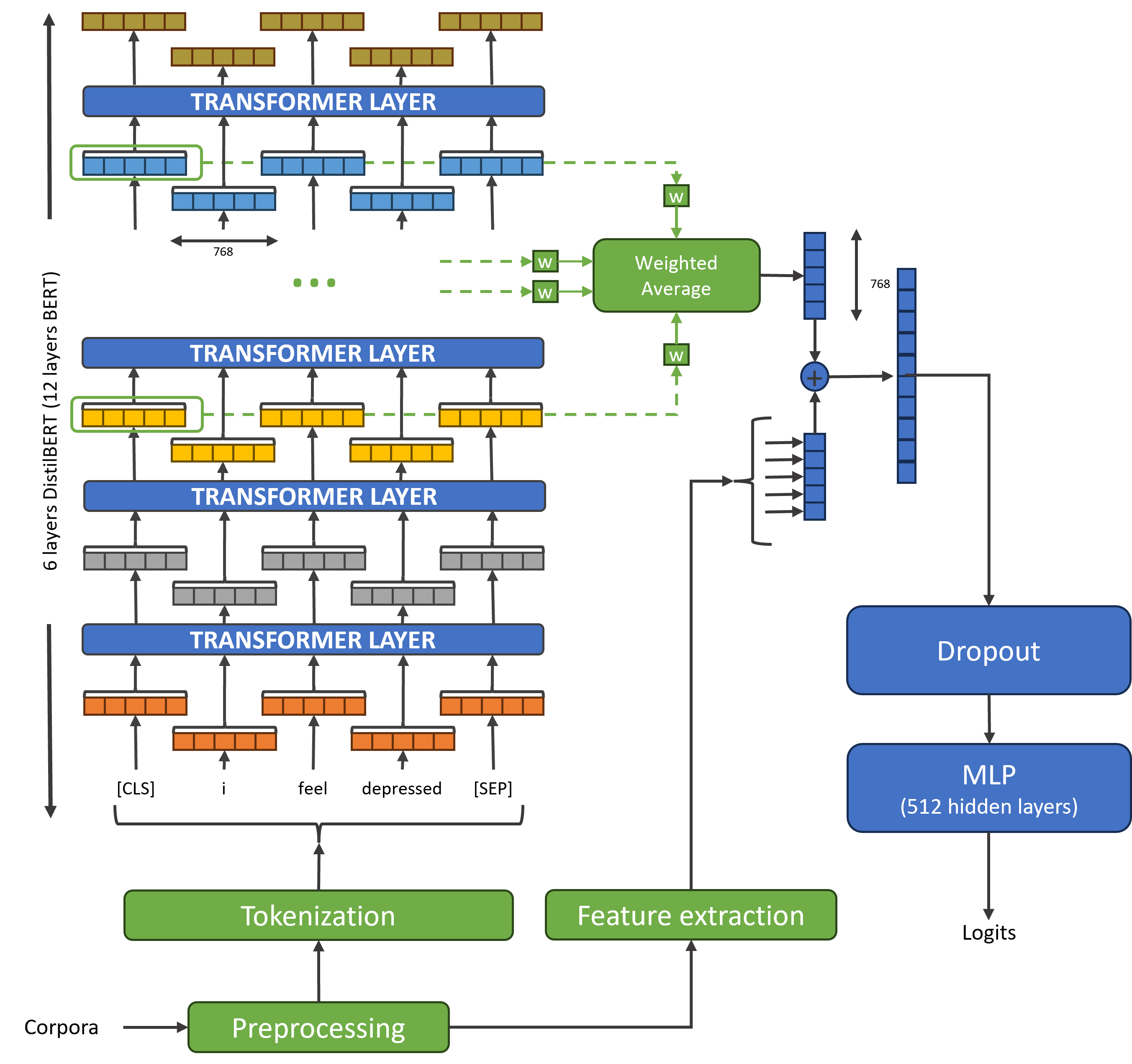}
    \caption{The proposed methodology}
    \label{fig:model_en}
\end{figure*}

Our research methodology (Figure \ref{fig:model_en}) is based on a hybrid model that combines a transformer model, specifically the Distilbert-base-uncased model, with a Multilayer Perceptron (MLP). The Distilbert-base-uncased model is a pre-trained transformer model that has been trained on a large corpus of English data in a self-supervised fashion. This model is uncased, meaning it does not differentiate between upper- and lower-case English letters.

The first step in our methodology involves tokenizing the input text. Tokenization is the process of converting the input text into a sequence of tokens or words. The tokenized text is then passed to the transformer model.

In our architecture, we use the hidden states from the last four layers of the transformer. These layers are chosen because deeper layers in the BERT model capture more complex semantic information \cite{song_utilizing_2020}. We compute a weighted average of these hidden states, with the weights learned during training. This allows the model to decide the importance of the output from each layer. The output of this process is a tensor that represents the weighted average hidden states of the transformer model. Specifically, we use the CLS tokens, which are the first tokens in each sequence and are used to aggregate the representation of the entire sequence. The transformer accepts an input of 512 tokens and has a hidden size of 768, meaning the dimension of the hidden states is 768.

The averaged CLS token is then combined with additional features and metadata to form the input for the MLP.  The metadata are the features extracted as described in Section \ref{subsec:feature_extr}. Before the input is passed to the MLP, it is subjected to a dropout layer with a dropout rate of 0.1 for regularization. This helps prevent overfitting by randomly setting a fraction of input units to 0 at each update during training time.

In our hybrid model, the classification head plays a crucial role in transforming the rich contextualized representations obtained from the transformer layers into actionable predictions. Positioned at the end of the architecture, it consists of a simple feed-forward neural network, the Multi-Layer Perceptron (MLP). The MLP, serving as the classification head of the transformer, is composed of a linear layer, a ReLU activation function, and another linear layer. The input to the MLP comprises the combined features, including the weighted average hidden states from the Distilbert-base-uncased model, along with additional features and metadata.

Tailoring the versatile information encoded by the transformer, the classification head acts as the decision-making module, ensuring adaptability and effectiveness across diverse classification scenarios. The hidden size of the MLP is set to 512, aligning with the design choices made for the transformer layers. The output dimension of the MLP is determined by the number of labels in the targeted task, ensuring the model's capacity to generate accurate predictions. In our case, the task involves four labels, reflecting the specific requirements of the classification task at hand.

The output of the MLP is the final output of the hybrid model (see Figure~\ref{fig:model_en}),  with model size of 66 million trainable parameters (66,774,536). This architecture allows us to leverage the power of the transformer for representation learning, while also incorporating additional features and metadata and a custom MLP for the specific task. The use of a weighted average of the hidden states from the Distilbert-base-uncased model allows the model to effectively utilize the representations learned at different layers of the transformer model. This, combined with the ability to incorporate additional features and metadata and a custom MLP, makes this architecture flexible and powerful for a variety of tasks.

The weighted average of hidden states from the transformer can be represented as follows:
\[
H_{\text{avg}} = \sum_{i=1}^{4} \alpha_i H_i
\]

Here, \(H_{\text{avg}}\) represents the weighted average hidden states, \(H_i\) represents the hidden states from each of the last four layers, and \(\alpha_i\) represents the learned weights for each layer.

The input to the MLP can be represented as follows:
\[
\text{MLP\_input} = [\text{CLS\_avg} : \text{additional\_features}]
\]

Where \(\text{CLS\_avg}\) is the averaged CLS token output from the transformer, and \(\text{additional\_features}\) represent the additional features and metadata.

The output of the MLP can be represented as follows:
\[
\text{MLP\_output} = \text{MLP}(\text{MLP\_input})
\]

Where \(\text{MLP\_output}\) represents the final output of the hybrid model, and \(\text{MLP}\) denotes the operations performed by the Multilayer Perceptron.

\section{Experiments}
\label{sec:exp}
\subsection{Baselines}
In our study, we compare our proposed methodology with several baseline models. 

\begin{itemize}
\item The first baseline is the approach presented by Ilias et. al. \cite{ilias_calibration_2023}. This baseline model is a transformer-based approach that integrates additional linguistic information into the BERT and MentalBERT models. The method begins by extracting a variety of linguistic features, including NRC Sentiment Lexicon, features derived by Latent Dirichlet Allocation (LDA) topics, Top2vec, and Linguistic Inquiry and Word Count (LIWC) features. These features are then projected to the same dimensionality as the outputs of the transformer models. The authors concatenate the representations obtained by BERT (or MentalBERT) and the linguistic information, applying a Multimodal Adaptation Gate to control the importance of each representation. The combined embeddings are then passed through a BERT (or MentalBERT) model, with the classification [CLS] token fed to Dense layers to produce the final prediction. To prevent the models from becoming overly confident, the authors employ label smoothing. They tested their proposed approaches on two publicly available datasets, which differentiate stressful from non-stressful texts, and posts indicating varying severity of depression (minimal, mild, moderate, and severe), which is the one we use.

\item The second baseline model comes from the paper by Yang et al. \cite{yang_mental_2022}. In this approach, the authors utilized BERT and MentalBERT for depression detection.

\item The third baseline involves the use of ALBERT with an LSTM layer. This baseline method has been used in \cite{naseem_early_2022}. ALBERT (A Lite BERT) is a more efficient version of BERT that reduces the model's size and training time while maintaining performance. In this baseline, the ALBERT model is combined with an LSTM layer to capture sequential dependencies within the text data. The final hidden state from the LSTM layer is used as the representation for classification.

\item The fourth baseline is the ALBERT model \cite{lan_albert_2019} paired with a Bidirectional LSTM (BiLSTM) layer. This baseline method has been used in \cite{naseem_early_2022}. This model allows the processing of text data in both forward and backward directions, capturing more contextual information than a standard LSTM. The output representations from both directions are concatenated and passed through dense layers for the final classification.

\item Another baseline utilizes Linguistic Inquiry and Word Count (LIWC) \cite{pennebaker_LIWC_2015} features combined with a Random Forest (RF) classifier. LIWC is a widely used tool for psycholinguistic analysis, which provides insights into the emotional, cognitive, and structural components of the text. These features are then input into a Random Forest classifier, which operates by constructing multiple decision trees and outputting the mode of the classes predicted by individual trees.

\item Finally, we consider a baseline using MentalBERT \cite{ji_mentalbert_2022} with an LSTM layer. MentalBERT is a domain-specific version of BERT, fine-tuned on mental health-related corpora, making it particularly effective for tasks related to mental health. In this baseline, the MentalBERT model's outputs are passed through an LSTM layer, and the final hidden state is used for the classification task.

\end{itemize}


\subsection{Experimental Setup}
For the experimental setup, the neural network model was implemented in a Python notebook on Google Colab and Kaggle, utilizing prominent libraries such as Hugging Face's Transformers\cite{wolf-etal-2020-transformers}, PyTorch\cite{NEURIPS2019_9015} for model construction and nlpaug\cite{ma2019nlpaug} for text augmentation. The evaluation of the model's performance was facilitated by leveraging the comprehensive metrics available in the scikit-learn library\cite{scikit-learn}. To accelerate the training process, a T4 GPU on Google Collab and a P100 GPU on Kaggle were employed, optimizing the computational efficiency. The training phase was conducted with a batch size of 8 and was iterated over 11 epochs, while 80\% of the dataset was utilized for training. We train and test our proposed approach five times and report the mean and standard deviation over five runs. For the optimization process, the Cross-Entropy Loss function was adopted, and the Adam optimizer with a learning rate of 1e-5 was utilized to fine-tune the model's parameters effectively. A list of hyperparameters is presented in Table~\ref{hyperparams}. This experimental setup allowed for the thorough evaluation and classification of depressive posts, showcasing the effectiveness of the proposed neural network in addressing the task at hand.

\begin{table}[!ht]
    \centering
    \caption{Hyperparameter values}
    \begin{tabular}{|c|c|}
    \hline
        \textbf{Hyperparameter} & \textbf{Value} \\ \hline
        Epochs & 11 \\ \hline
        Batch size & 8 \\ \hline
        Train size & 0.8 \\ \hline
        Optimizer & Adam \\ \hline
        DistilBert dimension & 768 \\ \hline
        Number of Features & 31 \\ \hline
        MLP layers &  3 (Linear, ReLU, Linear)\\ \hline
        MLP hidden size & 512 \\ \hline
        Number of layers to concatenate & 4 \\ \hline
        Learning rate of Adam optimizer & 0.00001 \\ \hline
        Dropout rate  & 0.1 \\ \hline
    \end{tabular}
    \label{hyperparams}
\end{table}

\subsection{Evaluation Metrics}
For the evaluation of the proposed neural network, we employed precision, recall, and F1-score as our primary metrics, all of which were weighted to account for class imbalances in the dataset. These metrics are widely recognized and commonly used in natural language processing tasks, including sentiment analysis and text classification. The utilization of weighted metrics allowed us to appropriately assess the model's performance in handling imbalanced classes, particularly in identifying depressive posts accurately. The choice of these evaluation metrics aligns with the approach used by Ilias et al. \cite{ilias_calibration_2023} By adopting similar evaluation metrics, we aimed to facilitate comparisons and ensure the reproducibility of results across studies in the field of depression detection in social media content.

\section{Results}
\label{sec:results}
In the course of our experiments, we trained and evaluated multiple configurations of our models. We observed that the classification head MLP model with 512 hidden layers and 'distilbert-base-uncased' outperformed the other configurations.  This model achieved a weighted precision of 84.26\%, a weighted recall of 84.18\%, and a weighted F1-score of 84.15\%. The classification head MLP model with 512 hidden layers using the 'distilbert-base-uncased' achieved an improvement of approximately 23.86\% in the weighted F1-score over the worst-performing baseline model and an improvement of approximately 10.99\% over the best-performing baseline model. A comprehensive comparison of our model's performance against the baseline models is presented in Table \ref{tab:performance_comparison}.

In comparison to the baseline models proposed by Ilias et.al. \cite{ilias_calibration_2023} and Yang et al.\cite{yang_mental_2022}, our best-performing model yielded competitive results. Ilias's best model, M-MentalBERT (LDA topics), showed a weighted precision of 73.74\%, a weighted recall of 73.23\%, and a weighted F1-score of 73.16\%. Yang et al.'s BERT model yielded a weighted precision of 72.99\%, a weighted recall of 71.97\%, and a weighted F1-score of 71.00\%, while their MentalBERT model showed a weighted precision of 73.35\%, a weighted recall of 70.81\%, and a weighted F1-score of 71.67\%. In contrast, Ilias's worst model, M-BERT (NRC), showed a weighted precision of 74.48\%, a weighted recall of 70.08\%, and a weighted F1-score of 69.96\%. We further benchmarked our best-performing model against additional baseline models from other studies, as detailed in Table \ref{tab:performance_comparison}. These baselines include ALBERT-based architectures combined with LSTM and BiLSTM layers, as well as models leveraging LIWC features with random forest, and a MentalBERT model combined with LSTM. Importantly, we ran these baseline models on our own dataset to ensure a consistent and fair comparison. The results show that our model, the classification head MLP with 512 hidden layers using 'distilbert-base-uncased,' significantly outperforms these baselines across all metrics. Specifically, our model achieved a weighted precision of 84.26\%, a weighted recall of 84.18\%, and a weighted F1-score of 84.15\%. In contrast, the best-performing baseline in this group, ALBERT + BiLSTM, achieved only 66.83\% in the weighted F1-score, indicating a substantial improvement of over 17.32\% points with our approach. This further emphasizes the robustness and effectiveness of our model in detecting depression severity compared to a broader range of existing methods.

\begin{figure*}[!hbt]
    \centering
    
    \label{fig:conf matrix}
    \begin{tikzpicture}
        \begin{axis}[
                colormap={bluewhite}{color=(white) rgb255=(90,96,191)},
                xlabel=Predicted,
                xlabel style={yshift=-30pt},
                ylabel=Actual,
                ylabel style={yshift=20pt},
                xticklabels={Minimum, Mild, Moderate, Severe}, 
                xtick={0,...,3}, 
                xtick style={draw=none},
                yticklabels={Minimum, Mild, Moderate, Severe}, 
                ytick={0,...,3}, 
                ytick style={draw=none},
                enlargelimits=false,
                colorbar,
                xticklabel style={
                  rotate=90
                },
                nodes near coords={\pgfmathprintnumber[fixed, precision=2]\pgfplotspointmeta},
                nodes near coords style={
                    yshift=-7pt
                },
            ]
            \addplot[
                matrix plot,
                mesh/cols=4, 
                point meta=explicit,draw=gray
            ] table [meta=C] {
                x y C
                0 0 0.8371
                1 0 0.0453
                2 0 0.0569
                3 0 0.05
                
                0 1 0.1884
                1 1 0.6299
                2 1 0.0715
                3 1 0.1
                
                0 2 0.2580
                1 2 0.06
                2 2 0.5975
                3 2 0.0780
        
                0 3 0.17
                1 3 0.0588
                2 3 0.0763
                3 3 0.6863
                
            }; 
        \end{axis}
    \end{tikzpicture}
    \caption{Confusion matrix of the best performing model on multi-class depression classification}
\end{figure*}
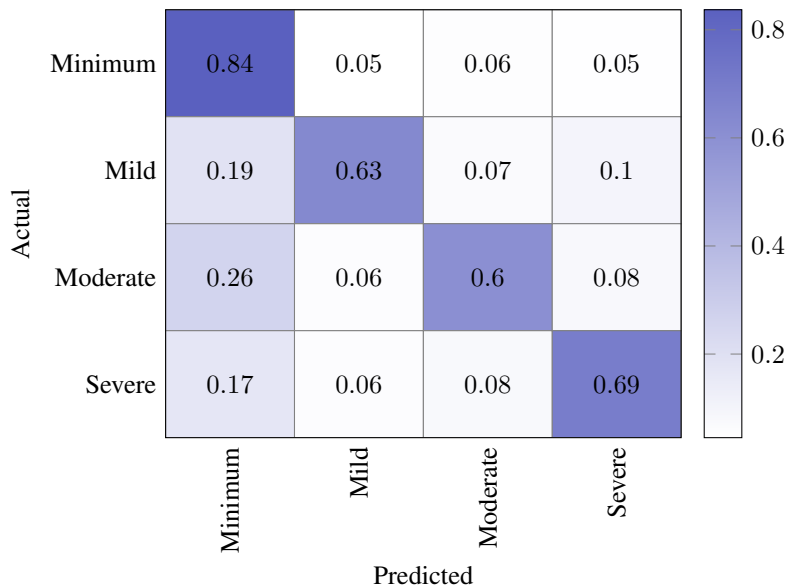

Lastly Figure \ref{fig:conf matrix} presents the confusion matrix for the classification model used to detect depression severity, detailing the performance across four classes: Minimum, Mild, Moderate, and Severe. The main diagonal cells indicate the model's accuracy for each class, with 84\% for Minimum, 63\% for Mild, 60\% for Moderate, and 69\% for Severe. The off-diagonal cells show the misclassification rates: Minimum is misclassified 5\% as Mild, 6\% as Moderate, and 5\% as Severe; Mild is misclassified 19\% as Minimum, 7\% as Moderate, and 10\% as Severe; Moderate is misclassified 26\% as Minimum, 6\% as Mild, and 8\% as Severe; Severe is misclassified 17\% as Minimum, 6\% as Mild, and 8\% as Moderate. This confusion matrix highlights the model's strengths and areas for improvement in distinguishing between different severity levels of depression.

\begin{table*}[!hbt]
\centering
\caption{Performance comparison of our best model (averaged over 5 runs and containing the standard deviation of the scores) and baseline models. The improvement of our model over the worst baseline model in terms of the weighted F1-score is approximately 14.19\%, and over the best baseline model is approximately 10.99\%.}
\label{tab:performance_comparison}
\resizebox{\textwidth}{!}{
\begin{tabular}{|l|l|l|l|}
\hline
\textbf{Model} & \textbf{Weighted Precision (\%)} & \textbf{Weighted Recall (\%)} & \textbf{Weighted F1-score (\%)} \\
\hline
M-MentalBERT - LDA topics \cite{ilias_calibration_2023} & 73.74\% & 73.23\% & 73.16\% \\
\hline
M-BERT - NRC \cite{ilias_calibration_2023} & 74.48\% & 70.08\% & 69.96\% \\
\hline
BERT \cite{yang_mental_2022} & 72.99\% & 71.97\% & 71.00\% \\
\hline
MentalBERT \cite{yang_mental_2022} & 73.35\% & 70.81\% & 71.67\% \\ \hline
ALBERT + LSTM & 62.87\% $\pm$ 4.83\% & 73.93\% $\pm$ 1.22\% & 65.37\% $\pm$ 2.21\% \\ \hline
ALBERT + BiLSTM & 64.83\% $\pm$ 1.42\% & 72.53\% $\pm$ 0.67\% & 66.83\% $\pm$ 0.55\% \\ \hline
 LIWC + Random forest & 53.50\% & 71.70\% & 60.29\% \\ \hline
MentalBERT+ LSTM  & 62.85\% $\pm$ 10.06\% & 72.87\% $\pm$ 0.28\% & 66.20\% $\pm$ 0.48\% \\
\hline
\hline
\makecell[l]{\textbf{Our: Classification Head MLP}\\ \textbf{(512 hidden layers, distilbert-base-uncased)}} & 84.26\% $\pm$ 0.48 \%	& 84.18\% $\pm$ 1.26\%	& 84.15\% $\pm$ 0.09\%\\
\hline
\end{tabular}
}
\end{table*}

\section{Ablation Study}
\label{sec:ablation}
In the process of model design and selection, we conducted an ablation study to ascertain the impact of the number of transformer layers, the type of transformer models, and the classification heads used in the architecture.

\subsection{Transformer Layers}

We explored a range of 1 to 6 transformer layers for our architecture with both MLP and LSTM classification layers. For the MLP classification head results showed that the model achieved its peak performance with four transformer layers, achieving a weighted  F1-score of 84.15\%. Both a decrease or an increase in the number of layers from this point resulted in a decrement in the F1-score, highlighting the importance of the right balance in model depth (see Tables \ref{table:ablation:layers_lstm} and \ref{table:ablation:layers_mlp}).

\begin{table*}[!hbt]
    \centering
    \caption{Ablation study for different layers with LSTM classification head and bert-base-uncased transformer.}
    \label{table:ablation:layers_lstm}
    \resizebox{\textwidth}{!}{
    \begin{tabular}{|c|c|c|c|c|}
    \hline
        \textbf{Classification Head and Hidden Size} & \textbf{number of transformer layers in use} & \textbf{Weighted Precision (\%)} & \textbf{Weighted Recall (\%)} & \textbf{Weighted Fl-score (\%)} \\ \hline
            LSTM 512 & 1 & 83.65\% & 83.70\% & 83.66\%\\ \hline
            LSTM 512 & 2 & 83.21\% & 83.35\% & 82.88\%\\ \hline
            LSTM 512 & 3 & 83.22\% & 83.47\% & 83.16\%\\ \hline
            LSTM 512 & 4 & 84.01\% & 82.43\% & 82.86\%\\ \hline
            LSTM 512 & 5 & 82.52\% & 82.89\% & 82.58\%\\ \hline
            LSTM 512 & 6 & 84.09\% & 84.27\% & 84.13\%\\ \hline
    \end{tabular}
    }
\end{table*}

\begin{table*}[!htb]
    \centering
    \caption{Ablation study for different layers with MLP classification head and distilbert-base-uncased transformer.}
    \label{table:ablation:layers_mlp}
    \resizebox{\textwidth}{!}{
    \begin{tabular}{|c|c|c|c|c|}
    \hline
        \textbf{Classification Head and Hidden Size} & \textbf{number of transfromer layers in use} & \textbf{Weighted Precission (\%)} & \textbf{Weighted Recall (\%)} & \textbf{Weighted Fl-score (\%)} \\ \hline
            MLP 512 & 1 & 81.73\% &  82.09\% &  81.52\% \\ \hline
            MLP 512 & 2 & 83.38\% &  82.43\% &  82.71\% \\ \hline
            MLP 512 & 3 & 81.86\% &  81.86\% &  81.52\% \\ \hline
            MLP 512 & 4 & 84.26\% &  84.18\% &  84.15\% \\ \hline
            MLP 512 & 5 & 82.46\% &  81.86\% &  81.49\% \\ \hline
            MLP 512 &   6 & 82.90\% &  82.89\% &  82.80\% \\ \hline
    \end{tabular}
    }
\end{table*}

\subsection{Choice of Transformer Models}

We compared the performance of two transformer models, BERT-base-uncased and DistilBERT-base-uncased. Interestingly, the study revealed no significant difference in performance between the two. Both models offered comparable weighted F1-scores, suggesting the choice of transformer model has little impact on the overall performance (see Table \ref{table:ablation:transformers}).

Nevertheless, given there was no noticeable difference in performance, we opted for DistilBERT-base-uncased as the transformer model in our architecture. This selection is driven by DistilBERT's characteristics of being lighter and faster than BERT, offering reduced computational costs, faster training and prediction times, and lower memory requirements. Such features make it a more feasible choice, especially for deployment in resource-constrained environments.

\begin{table}[!hbt]
    \centering
    \tiny
    \caption{Ablation study for different transformers.}
    \label{table:ablation:transformers}
    \begin{tabular}{|c|c|c|c|c|}
    \hline
        \makecell[c]{\textbf{Classification Head} \\ \textbf{and Hidden Size}} & \textbf{models} & \textbf{W. Precision (\%)} & \textbf{W. Recall (\%)} & \textbf{W. Fl-score (\%)} \\ \hline
        MLP 512	& bert-base-uncased &	84.37\% &84.27\% & 84.30\% \\ \hline
        MLP 512	& distilbert-base-uncased &	84.26\% &84.18\% & 84.15\% \\ \hline
        LSTM 512 & bert-base-uncased &	84.01\% &82.43\% & 82.86\% \\ \hline
        LSTM 512 & distilbert-base-uncased &	76.92\% & 75.38\% & 74.88\% \\ \hline
    \end{tabular}
\end{table}

\subsection{Classification Heads}
The classification head, a crucial component in the model, plays a pivotal role in processing and interpreting the features extracted by the preceding layers.

We investigated the performance of five different types of classification heads: MLP, LSTM, MM-Gate and MM-Xatt. The gate mechanisms are used in multiple state-of-the-art papers \cite{ilias_multimodal_2022,ilias_detecting_2023} and show promising results.

Our results showed that the MLP and LSTM classification heads delivered the highest performance, with MLP being the superior choice due to its simplicity and efficiency. The gating mechanisms, while novel, didn't contribute to a superior performance in our task (see Table \ref{table:ablation:head}).
\begin{table}[!hbt]
    \centering
    \tiny
    \label{table:ablation:head}
    \caption{Ablation study for different classification heads}
    \begin{tabular}{|c|c|c|c|}
    \hline
        \makecell[c]{\textbf{Classification Head} \\ \textbf{and Hidden Size}} & \textbf{Weighted Precision (\%)} & \textbf{Weighted Recall (\%)} & \textbf{Weighted Fl-score (\%)} \\ \hline
        MLP 512 &	84.26\% &84.18\% & 84.15\% \\ \hline
        LSTM 512 & 76.92\% & 75.38\% & 74.88\% \\ \hline
        MM-Gate 512 & 80.43\% & 80.14\% & 79.88\% \\ \hline
        MM-Xatt 512 & 51.00\% & 61.65\% & 54.71\% \\ \hline
    \end{tabular}
\end{table}

\subsection{Augmentation}

\begin{table*}[!hbt]
\centering
\tiny
\caption{Difference of scores before and after augmentation}
\label{tab:augmentation_scores}
\resizebox{\textwidth}{!}{
\begin{tabular}{|l|l|l|lll|lll|lll|}
\hline
                                                                                                    &                                                                                                         &                                                                                 & \multicolumn{3}{c|}{\begin{tabular}[c]{@{}c@{}}\textbf{Before Augmentation}\\ \textbf{(Weighted Scores)}\end{tabular}} & \multicolumn{3}{c|}{\begin{tabular}[c]{@{}c@{}}\textbf{After Augmentation}\\ \textbf{(Weighted Scores)}\end{tabular}} & \multicolumn{3}{c|}{\textbf{Difference}}                                                                 \\ \hline
\multicolumn{1}{|c|}{\begin{tabular}[c]{@{}c@{}}\textbf{classification}\\ \textbf{head}\end{tabular}} & \multicolumn{1}{c|}{\begin{tabular}[c]{@{}c@{}}\textbf{Number}\\ \textbf{of trans.} \\ \textbf{layers} \\ \textbf{in   use}\end{tabular}} & \multicolumn{1}{c|}{\begin{tabular}[c]{@{}c@{}}\textbf{models}\\ \textbf{(uncased)}\end{tabular}} & \multicolumn{1}{c|}{\textbf{Prec.}}   & \multicolumn{1}{c|}{\textbf{Rec.}}    & \multicolumn{1}{c|}{\textbf{Fl-score}}   & \multicolumn{1}{c|}{\textbf{Prec.}}   & \multicolumn{1}{c|}{\textbf{Rec.}}    & \multicolumn{1}{c|}{\textbf{Fl-score}}  & \multicolumn{1}{c|}{\textbf{Prec.}} & \multicolumn{1}{c|}{\textbf{Rec.}}   & \multicolumn{1}{c|}{\textbf{Fl-score}} \\ \hline
LSTM                                                                                            & 1                                                                                                       & bert                                                               & \multicolumn{1}{l|}{71.22\%}      & \multicolumn{1}{l|}{73.41\%}   & 71.96\%                         & \multicolumn{1}{l|}{83.65\%}      & \multicolumn{1}{l|}{83.70\%}   & 83.66\%                        & \multicolumn{1}{l|}{12.42\%}    & \multicolumn{1}{l|}{10.29\%}  & 11.69\%                       \\ \hline
LSTM                                                                                            & 2                                                                                                       & bert                                                               & \multicolumn{1}{l|}{74.38\%}      & \multicolumn{1}{l|}{74.54\%}   & 73.88\%                         & \multicolumn{1}{l|}{83.21\%}      & \multicolumn{1}{l|}{83.35\%}   & 82.88\%                        & \multicolumn{1}{l|}{8.83\%}     & \multicolumn{1}{l|}{8.81\%}   & 9.00\%                        \\ \hline
LSTM                                                                                            & 3                                                                                                       & bert                                                               & \multicolumn{1}{l|}{74.82\%}      & \multicolumn{1}{l|}{74.26\%}   & 74.44\%                         & \multicolumn{1}{l|}{83.22\%}      & \multicolumn{1}{l|}{83.47\%}   & 83.16\%                        & \multicolumn{1}{l|}{8.40\%}     & \multicolumn{1}{l|}{9.21\%}   & 8.72\%                        \\ \hline
LSTM                                                                                            & 4                                                                                                       & bert                                                               & \multicolumn{1}{l|}{74.67\%}      & \multicolumn{1}{l|}{72.98\%}   & 73.74\%                         & \multicolumn{1}{l|}{84.01\%}      & \multicolumn{1}{l|}{82.43\%}   & 82.86\%                        & \multicolumn{1}{l|}{9.33\%}     & \multicolumn{1}{l|}{9.45\%}   & 9.12\%                        \\ \hline
LSTM                                                                                            & 5                                                                                                       & bert                                                               & \multicolumn{1}{l|}{72.40\%}      & \multicolumn{1}{l|}{75.39\%}   & 72.86\%                         & \multicolumn{1}{l|}{82.52\%}      & \multicolumn{1}{l|}{82.89\%}   & 82.58\%                        & \multicolumn{1}{l|}{10.12\%}    & \multicolumn{1}{l|}{7.50\%}   & 9.72\%                        \\ \hline
LSTM                                                                                            & 6                                                                                                       & bert                                                               & \multicolumn{1}{l|}{72.86\%}      & \multicolumn{1}{l|}{71.00\%}   & 72.19\%                         & \multicolumn{1}{l|}{84.09\%}      & \multicolumn{1}{l|}{84.27\%}   & 84.13\%                        & \multicolumn{1}{l|}{11.23\%}    & \multicolumn{1}{l|}{13.27\%}  & 11.94\%                       \\ \hline
LSTM                                                                                            & 4                                                                                                       & distilbert                                                         & \multicolumn{1}{l|}{73.56\%}      & \multicolumn{1}{l|}{74.68\%}   & 73.75\%                         & \multicolumn{1}{l|}{76.92\%}      & \multicolumn{1}{l|}{75.38\%}   & 74.88\%                        & \multicolumn{1}{l|}{3.36\%}     & \multicolumn{1}{l|}{0.70\%}   & 1.13\%                        \\ \hline
MLP                                                                                             & 4                                                                                                       & bert                                                               & \multicolumn{1}{l|}{72.52\%}      & \multicolumn{1}{l|}{74.12\%}   & 73.16\%                         & \multicolumn{1}{l|}{84.37\%}      & \multicolumn{1}{l|}{84.27\%}   & \textbf{84.30\%}                        & \multicolumn{1}{l|}{11.84\%}    & \multicolumn{1}{l|}{10.15\%}  & 11.14\%                       \\ \hline
MLP                                                                                             & 1                                                                                                       & distilbert                                                         & \multicolumn{1}{l|}{73.86\%}      & \multicolumn{1}{l|}{75.81\%}   & \textbf{74.64\%}                         & \multicolumn{1}{l|}{81.73\%}      & \multicolumn{1}{l|}{82.09\%}   & 81.52\%                        & \multicolumn{1}{l|}{7.87\%}     & \multicolumn{1}{l|}{6.28\%}   & 6.88\%                        \\ \hline
MLP                                                                                             & 2                                                                                                       & distilbert                                                         & \multicolumn{1}{l|}{72.24\%}      & \multicolumn{1}{l|}{67.47\%}   & 69.30\%                         & \multicolumn{1}{l|}{83.38\%}      & \multicolumn{1}{l|}{82.43\%}   & 82.71\%                        & \multicolumn{1}{l|}{11.14\%}    & \multicolumn{1}{l|}{14.97\%}  & 13.42\%                       \\ \hline
MLP                                                                                             & 3                                                                                                       & distilbert                                                         & \multicolumn{1}{l|}{75.69\%}      & \multicolumn{1}{l|}{72.28\%}   & 72.73\%                         & \multicolumn{1}{l|}{81.86\%}      & \multicolumn{1}{l|}{81.86\%}   & 81.52\%                        & \multicolumn{1}{l|}{6.17\%}     & \multicolumn{1}{l|}{9.58\%}   & 8.79\%                        \\ \hline
MLP                                                                                             & 4                                                                                                       & distilbert                                                         & \multicolumn{1}{l|}{71.99\%}      & \multicolumn{1}{l|}{73.41\%}   & 72.59\%                         & \multicolumn{1}{l|}{84.26\%}      & \multicolumn{1}{l|}{84.18\%}   & 84.15\%                        & \multicolumn{1}{l|}{12.27\%}    & \multicolumn{1}{l|}{10.77\%}  & 11.56\%                       \\ \hline
MLP                                                                                             & 5                                                                                                       & distilbert                                                         & \multicolumn{1}{l|}{71.67\%}      & \multicolumn{1}{l|}{74.82\%}   & 72.78\%                         & \multicolumn{1}{l|}{82.46\%}      & \multicolumn{1}{l|}{81.86\%}   & 81.49\%                        & \multicolumn{1}{l|}{10.79\%}    & \multicolumn{1}{l|}{7.04\%}   & 8.71\%                        \\ \hline
MLP                                                                                             & 6                                                                                                       & distilbert                                                         & \multicolumn{1}{l|}{73.88\%}      & \multicolumn{1}{l|}{74.68\%}   & 74.22\%                         & \multicolumn{1}{l|}{82.90\%}      & \multicolumn{1}{l|}{82.89\%}   & 82.80\%                        & \multicolumn{1}{l|}{9.02\%}     & \multicolumn{1}{l|}{8.21\%}   & 8.58\%                        \\ \hline
MM-Gate                                                                                         & 4                                                                                                       & distilbert                                                         & \multicolumn{1}{l|}{73.25\%}      & \multicolumn{1}{l|}{71.57\%}   & 71.95\%                         & \multicolumn{1}{l|}{80.43\%}      & \multicolumn{1}{l|}{80.14\%}   & 79.88\%                        & \multicolumn{1}{l|}{7.18\%}     & \multicolumn{1}{l|}{8.57\%}   & 7.93\%                        \\ \hline
MM-Xatt                                                                                         & 4                                                                                                       & distilbert                                                         & \multicolumn{1}{l|}{58.16\%}      & \multicolumn{1}{l|}{72.98\%}   & 63.54\%                         & \multicolumn{1}{l|}{1.77\%}       & \multicolumn{1}{l|}{13.32\%}   & 3.13\%                         & \multicolumn{1}{l|}{-56.39\%}   & \multicolumn{1}{l|}{-59.66\%} & -60.41\%                      \\ \hline
\end{tabular}
}
\end{table*}

Table \ref{tab:augmentation_scores} presents the difference in weighted scores for various machine learning models before and after applying data augmentation techniques. When considering the LSTM model with one layer and BERT (uncased), the Precision increased from 71.22\% to 83.65\%, Recall from 73.41\% to 83.70\%, and F1-score from 71.96\% to 83.66\%. In a similar LSTM configuration but with 6 layers, the F1-score showed a significant increase from 72.19\% to 84.13\%. The difference in the F1-score ranged between 8\% to nearly 12\% for different LSTM configurations with BERT. However, for LSTM with 4 layers using DistilBERT, the change in the F1-score was minimal at just 1.13\%. In contrast, using an MLP architecture, improvements were also noted but to varying degrees. For instance, with 4 layers and BERT, the F1-score went up from 73.16\% to 84.30\%. Interestingly, when using DistilBERT and 6 layers, the F1-score improved from 74.22\% to 82.80\%. The MM-Gate model with 4 layers and DistilBERT also saw an increase in the F1-score from 71.95\% to 79.88\%. However, the MM-Xatt configuration with 4 layers and DistilBERT notably decreased in it's performance metrics, with the F1-score dropping dramatically from 63.54\% to just 3.13\%. Overall, the data shows that while data augmentation generally improved the performance of LSTM and MLP architectures, the benefits were not universally observed across all configurations and models.

\section{Qualitative and Error Analysis}

In depression detection on social media, mismatches in predictions can arise due to several factors, including imbalanced data where certain labels are underrepresented. For example, the minimum label might have significantly more posts than others, leading to an over-representation of less severe cases in the dataset. This imbalance can cause models to lean toward predicting posts as minimum or moderate even when they describe more severe symptoms. Another common issue is the similarity between posts assigned to different labels, leading to false predictions. Below are examples that highlight this phenomenon:

\textbf{Example 1:} Actual Label: Severe, Predicted Label: Moderate

In the first post, which is actually labeled as moderate, the individual discusses continuous panic attacks, intrusive thoughts, and constant fear:

\textit{i have panic attack after panic attack and... i truly just don\'t know what to do anymore. no ono matter how i explain the severity of my situation to people, but... still somehow they seem to not understand just how miserable my mind makes me everyday. i get intrusive thoughts like about how something terrible could happen to my family, or a lot of " what... ifs ". my mind is running 24 / 7 and it\'s driving everyone insane. i have a terrible fear of danger along with an even more crippling risk of had something happening happened to dear friends ( my parents, brothers.. ) all day and my mind continually plays over thoughts and scenarios that leave me sad, scared, constantly wired with fear, and never... all exhausted.}

This post shares many similarities with the one below, which is actually labeled as severe but was mismatched as moderate. Both posts describe intense anxiety and intrusive thoughts. However, the second post goes deeper into feelings of isolation and overwhelming distress, which makes the misclassification notable:

\textit{i am slightly introverted and however the broken friendship i had since my previous roommate made me feel very anxious, so i settled into an apartment on my own. my am living alone, yet when i have severe anxiety attacks it becomes lonely and seems almost unbearable. i sometimes do think so much that it feels like i am going completely crazy. i always have terrifying thoughts and often i make up scenarios in which i have undergone some terrible condition, or when i am dying. her parents know instinctively that a have anxiety, yes but never took it more seriously.}

\textbf{Example 2:} Actual Label: Minimum, Predicted Label: Moderate

In the first post, which is actually labeled as moderate, the individual goes deeper into trauma-related anxiety and its impact on intimate relationships:

\textit{I don’t remember a lot of it I just remember little snip bits of what happened. I don’t even remember if there was penetration and I’m hoping there wasn’t. I still have problems trying to remember what happened and I feel like if I remember it all I’m going to have a breakdown. Now that I’m older I’m starting to have sex and be intimate with others. I’m noticing a pattern where my body is like rejecting my partner and I’m concerned it might be caused by my abuse at an early age.}

This post expresses ongoing anxiety and emotional turmoil, which makes it similar to the post below. However, the following post, labeled as minimum but predicted as moderate, discusses the same trauma but in a less intense way, showing the mismatch:

\textit{I’m noticing a pattern where my body is like rejecting my partner and I’m concerned it might be caused by my abuse at an early age. Should I seek counseling? But I’m afraid if I do I’m going to have to talk more about what happened and I’m going to break. I’ve talked to therapist before but whenever the topic of the abuse arises I tense up and can’t remember anything. I’m sorry if I’m rambling on it’s just a hard subject for me to talk about and I don’t know how to put into words the emotions I feel towards these events.}

\textbf{Example 3:} Actual Label: Mild, Predicted Label: Moderate

In the first post, which is actually labeled as moderate, the individual describes persistent anxiety and how it disrupts daily life:

\textit{As of Sunday i have experience sever anxiety. Its too the point i am unable to do normal everyday functions and doing my job is harder. All i can think about is how i was ripped away from my daughter and how she is without me. Its to the point my therapist uas suggested i used cbd oil to manage my anxiety instead of narcotics. They help take the edge off and let me relax to the point i can sleep.}

This post is very similar to the one below, where the person describes episodic spikes of anxiety but has general control over it. The model likely predicted the following post as moderate due to these similarities, despite its actual mild classification:

\textit{Normally, my anxiety is very well controlled. I meditate every morning for \~15 minutes and have been in therapy for the better part of the last 3 or 4 years. I feel WAY better than I used to, and on a day to day basis things are great. Buttt every once in a while (A handful of times a year, tops) something will realllly set me over the edge, and send me into an intense anxiety spiral where I compulsively ask 5 or so different friends for advice on what to do, post a lot of threads online about what I should do, and ruminate on the topic for days or weeks. Sometimes I'll have chats in messenger about whatever it is that will draw out over an entire 3-4 hour period.}

\textbf{Example 4:} Actual Label: Moderate, Predicted Label: Severe

In the first post, which is labeled as severe, the person describes ongoing abusive experiences and the emotional toll they have taken:

\textit{he still forcefully holds me back in life. although he repeatedly still finds ways rann has finally fit inside me. eventually i will talk with him. and i always feel very soon after i do. i'and ve tried blocking him in our social media, but he still finds ways to even get use to me.}

This post shares many of the same abusive themes as the one below, which was labeled as moderate but was mismatched as severe due to the intensity of the described events, even though the emotional toll was less pronounced:

\textit{he has made most me almost die until i pretty literally threw something up, nearly forced me to eat my vomit. he plays computer games that are pure torture. so he knows how to break me down mentally until eventually i somehow just might become ruined over a period of their time. he also point guns at me. he made me play russian orthodox roulette ( turns out the gun wasn't loaded but he used a hand switch to make you appear loaded.}

 \section{Discussion}

\subsection{Limitations}

However, this study comes with some limitations. Specifically, we did not perform hyperparameter tuning due to limited access to GPU resources. It is known that hyperparameter tuning is a procedure yielding to higher evaluation results. Additionally, this study requires access to labelled datasets. Access to labelled datasets is often a difficult task. For this reason, methods including self-supervised learning have been developed aiming to address the problem of labels' scarcity. Additionally, we tested our approach only on one dataset. Finally, this study did not include an explainable approach, which will explain why this transformer-based network reached a specific decision. 

\subsection{Findings}

From the results of this study, we found that:

\begin{itemize}
    \item \textit{Finding 1:} We found that the data augmentation method increased the evaluation performance of the proposed model. Specifically, Weighted Precision, Recall, and F1-score presented a surge of 12.26\%, 10.75\%, and 11.52\% respectively.
    \item \textit{Finding 2:} Results showed that the introduced approach outperformed competitive baselines in weighted Precision by 9.78-11.27\%, weighted Recall by 10.95-14.10\%, and weighted F1-score by 10.99-14.19\%.
    \item \textit{Finding 3:} Findings from a series of ablation studies showed the effectiveness of the introduced architectures.
\end{itemize}

\subsection{Generalization to other social media platforms}

Our introduced approach can be easily adapted to other social media platforms. This can be justified by the fact that our proposed approach uses only posts on Reddit. Thus, one can use very easily posts on other social media platforms, including X (Twitter), Meta (Facebook), etc., and employ our introduced approach. 

However, it must be noted that the access of data to other social media platforms is a challenging task. For instance, Facebook does not easily share data with researchers \cite{yang2024social}. Additionally, Twitter has decided to end free access to their API. Specifically, the paid plans provide very limited data. For instance, social media researchers \cite{dang_elon_musk} were obliged to cancel, suspend, or modify more than 100 studies about Twitter. 

\subsection{Challenges of deploying the model in real-world settings}

Deploying our introduced model in real-world settings entails significant challenges. Although we have used DistilBERT, which is the distilled version of BERT and appears to be smaller, faster, cheaper, and lighter than BERT, its deployment still demands a lot of computational resources. Our approach has been developed for mental health monitoring. Thus, it needs to respond quickly in real-time applications processing at the same time huge amount of data. However, latency issues may arise. At the same time, maintaining the infrastructure, such as GPUs, cloud servers, etc., required for training and storing the AI models is a costly procedure. Another significant challenge arisen by the deployment in a real-world framework is the loss of accuracy due to different data. Specifically, our proposed model needs to be trained continuously so as to adapt to different data. Additionally, ensuring that the proposed approach performs equally well across different demographics or languages can be challenging. Deploying the model without thorough fairness testing could lead to biased outcomes. Finally, the deployment in real-world frameworks entails security issues. Specifically, machine learning models are vulnerable to adversarial attacks. Thus, robust AI models must be designed.

\section{Conclusion}
\label{sec:conclusion}
In this study, we addressed the critical task of early identification of depression severity levels in Reddit posts, leveraging a comprehensive approach that combines natural language processing techniques, deep learning, and data augmentation. Our research explored the limitations commonly observed in existing methodologies, such as overfitting, small dataset sizes, annotator biases, and the lack of age and gender awareness, and proposed a novel model to overcome some of these challenges.

The proposed model offers several advancements over the methodologies reported in the literature. By leveraging the power of BERT, a widely used transformer model, we extracted detailed linguistic patterns from the input text. This transformer-based approach overcame the limitations of traditional machine learning algorithms, offering a more sophisticated method for interpreting the complexities of natural language. Additionally, our model did not solely rely on transformer-based contextual embeddings but also incorporated auxiliary features, including metadata and linguistic markers. This strategic fusion enhanced the model's ability to comprehend each post holistically, capturing crucial nuances that might otherwise be overlooked. Consequently, our model enabled a more accurate identification of potential signs of depression.

Through extensive experiments, we demonstrated that our proposed model achieved competitive results compared to state-of-the-art baseline models. In particular, our model outperformed the best-performing baseline by approximately 10.94\% in terms of weighted F1-score. These results highlight the effectiveness of our approach in handling the nuanced task of depression severity level classification in social media content.

In our ablation study, we examined the impact of various model configurations, including the number of transformer layers, the choice of transformer models, classification heads and the use of data augmentation. These experiments allowed us to fine-tune our architecture and make informed decisions, ultimately leading to the selection of a model configuration that delivered the best performance.

Furthermore, we applied data augmentation techniques to address the challenges posed by class imbalance and dataset size. While data augmentation generally improved the performance of LSTM and MLP architectures, the benefits were not universally observed across all configurations and models. Nevertheless, our research contributes to the growing body of knowledge in the field of depression detection in social media content by offering a robust and comprehensive model that overcomes limitations and delivers more accurate results.

In conclusion, our study provides valuable insights into the early identification of depression severity levels in social media posts, with the proposed model showcasing significant advancements in performance. As the field continues to evolve, the contributions made here serve as a foundation for further research in the critical area of mental health awareness and early intervention through natural language processing and deep learning techniques.  Moreover, the potential impact of our study extends to mental health policy makers. By utilizing our model for fast monitoring of social media posts, it is possible to gauge the mental health of users in real-time, thereby enabling more timely and targeted interventions. This capability could transform how mental health crises are identified and managed, leading to better outcomes for individuals and communities. Finally, this study could be used as a tool for epidemiological research. Analyzing social media data for identifying severity of depression could assist public health authorities identify regions, age groups, or demographics particularly affected by depression. Epidemiological research constitutes a vital tool for improving public health outcomes, developing preventive strategies, and guiding healthcare policy.

\section{Future Directions}
\label{sec:future}
Delving into the interaction between linguistic cues and human emotions, this paper has paved the way for the detection of signs of depression in social media posts. While significant progress has been made, this work opens avenues for further exploration:

\begin{enumerate}
    \item \textbf{Using Different Datasets:} Social media, with its vast global reach, provides a rich source of data. Utilizing more extensive and diverse datasets can enhance the model's universality and sensitivity to various sociolinguistic nuances. For instance, we aim to use the Mental Health Blog Dataset, which includes posts written in English \cite{boinepelli-etal-2022-leveraging}. The publicly available datasets presented in \cite{pirina-coltekin-2018-identifying,ijcai2017p536} can be exploited. Datasets collected through the Covid-19 pandemic can also be used \cite{10241281, 10041797}. Our approaches could also be evaluated in datasets of the \textit{eRisk} depression severity estimation tasks \cite{10.1007/978-3-030-28577-7_27,10.1007/978-3-030-45442-5_72,10.1007/978-3-030-85251-1_22}. 

    \item \textbf{Merging Gate Mechanisms with Neural Classification Networks:} Combining gate functions with neural classification networks such as LSTM or CNN can capture richer linguistic insights. This hybrid approach could more effectively uncover complex patterns indicative of depressive tendencies.

    \item \textbf{Employing Larger and More Complex Transformers:} Leveraging state-of-the-art models like Chat GPT and Llama 2 could boost our model's accuracy. These transformers, with their advanced text processing capabilities, may reveal subtle signs of depression often overlooked.

    \item \textbf{Multi-dimensional Feature Analysis:} Beyond the main text, features like posting timestamps can offer additional insights. Tools like LIWC or POS tagging can further dissect posts, unraveling the embedded emotional and syntactic layers.

    \item \textbf{User History Integration:} Incorporating a user's posting history could provide information about the evolving mental state. This approach could be invaluable in distinguishing the cyclic or episodic nature of depressive tendencies, potentially aiding in early interventions.

    \item \textbf{Expanding to Multimodalities:} As multimedia content becomes integral to social media posts, our model could benefit from deciphering emotions embedded in images, videos, or links. Computer vision or content-based emotion analysis techniques can be applied to these expressions.



    \item \textbf{Broadening the Scope of Mental Health:} While our primary focus was depression, the framework could be adapted to detect other mental health conditions expressed on social media, from anxiety to PTSD or even anorexia.

    \item \textbf{Explainability:} To increase model explainability and understand its decisions better, we can explore the use of explainable AI methods \cite{ilias_explainable_2022} \cite{ilias_explainable_2022-1}. These methods can provide additional insights into how the model understands and categorizes emotions in posts. Integrating these methods into our model can aid researchers and mental health professionals in better comprehending the results and predictions related to depression and other mental health conditions on social media.


\end{enumerate}

In this interdisciplinary landscape of linguistics, technology, and mental well-being, the current research holds great promise. The horizon ahead is vast, inviting interdisciplinary collaboration and ongoing exploration, striving for a comprehensive understanding and support of the digital reflections of the human psyche.

\bibliographystyle{unsrt}
\bibliography{references}

\end{document}